\documentclass{article}

\PassOptionsToPackage{square,numbers}{natbib}
\usepackage[preprint]{neurips_2020}

\usepackage[utf8]{inputenc} 
\usepackage[T1]{fontenc}    
\usepackage{hyperref}       
\usepackage{url}            
\usepackage{booktabs}       
\usepackage{amsfonts}       
\usepackage{nicefrac}       
\usepackage{microtype}      
\usepackage[ruled,vlined]{algorithm2e}
\usepackage{graphicx, wrapfig}
\usepackage{amsmath}
\usepackage{subcaption}
\usepackage[labelfont=bf]{caption}
\usepackage{float}
\graphicspath{ {./figures/} }

\DeclareMathOperator*{\argmax}{arg\,max}

\newcommand{\beginsupplement}{%
        \setcounter{table}{0}
        \renewcommand{\thetable}{S\arabic{table}}%
        \setcounter{figure}{0}
        \renewcommand{\thefigure}{S\arabic{figure}}%
     }
     
\title{
Collective Learning by Ensembles of Altruistic Diversifying Neural Networks
}

\author{%
  Benjamin Brazowski \\
  Department of Neurobiology\\
  Weizmann Institute of Science\\
  Rehovot 76100, Israel \\
  \texttt{benjamin.brazowski@weizmann.ac.il} \\
   \And
   Elad Schneidman \\
  Department of Neurobiology\\
  Weizmann Institute of Science\\
  Rehovot 76100, Israel \\
   \texttt{elad.schneidman@weizmann.ac.il} \\
}

\begin{document}

\maketitle

\begin{abstract}
  
Combining the predictions of collections of neural networks often outperforms the best single network. Such ensembles are typically trained independently, and their superior `wisdom of the crowd' originates from the differences between networks. Collective foraging and decision making in socially interacting animal groups is often improved or even optimal thanks to local information sharing between conspecifics. We therefore present a model for co-learning by ensembles of interacting neural networks that aim to maximize their own performance but also their functional relations to other networks. We show that ensembles of interacting networks outperform independent ones, and that optimal ensemble performance is reached when the coupling between networks increases diversity and degrades the performance of individual networks. Thus, even without a global goal for the ensemble, optimal collective behavior emerges from local interactions between networks. We show the scaling of optimal coupling strength with ensemble size, and that networks in these ensembles specialize functionally and become more `confident’ in their assessments. Moreover, optimal co-learning networks differ structurally, relying on sparser activity, a wider range of synaptic weights, and higher firing rates - compared to independently trained networks. Finally, we explore interactions-based co-learning as a framework for expanding and boosting ensembles.
\end{abstract}

\section{Introduction}

Ensemble learning methods typically combine the outputs of several networks, models, or agents to obtain superior results than any of the individual models alone. Typically, the power of the ensemble arises from `error diversity' of the individual members of the ensemble, as in bootstrap methods such as bagging \cite{breiman_1996} and boosting \cite{freund_1996}. A simple and common approach is to train multiple networks that are initialized with different random parameters and combine them after training. The diversity between the ensemble members originates in this case from the random initial conditions and the stochastic nature of the learning algorithm.

Ensemble learning has mainly focused on training individual models independently from one another, and then merging them together \cite{webb_2019}. While even a simple averaging of their outputs or their parameters can be beneficial \cite{izmailov_2018}, more sophisticated `ensembling' can give even superior results \cite{huang_2017}. \citet{garipov_2018} explored the structure of the space of networks trained on the same task, in terms of their performance or loss function. They showed that between locally optimal deep networks (i.e. the sets of parameters that make a network superior to small changes in the parameters sets) there are simple `curves’ along which training and test accuracy are nearly constant. A training procedure to discover these high-accuracy pathways suggested a geometric way to merge variants of the optimal networks together to give high ensemble accuracy. Efficient ensembles may also be constructed by optimizing directly the performance of the ensemble: \citet{dutt_2019} showed that training of networks composed of sub-networks whose outputs were merged by either averaging their logits or multiplying their likelihood values outperformed the average accuracy of training the sub-networks independently. \citet{webb_2019} showed that the performance of a single large network was comparable to the performance of a collection of smaller sub-networks, with the same total number of hidden units, trained to optimize a global goal as well as local goals for each of its sub-networks.

Analysis and modeling of groups of insects, fish, birds, and mammals demonstrated the benefits of social interactions among group members, and often resulted in efficient or even optimal collective behavior \cite{ramdya_2015,strandburgpeshkin_2015,gelblum_2015}. While many models of collective behavior in animals rely on `mechanistic' couplings, these interactions imply that group members use their conspecifics as sources of information. Recent models of collective foraging \cite{karpas_2017} showed that when individual agents search for a source but also aim to increase their information diversity, efficient collective behavior emerges in groups of opportunistic agents, which is comparable to the optimal group behavior. Moreover, theoretical analysis of decision-making under natural physiological limitations has shown that optimal behavior can result from a collection of `selfish' agents that are in direct conflict with one another \cite{livnat_2006}. 

Inspired by models of collective behaviour in animals, artificial agents and ensemble learning, we introduce here a model for co-learning by ensembles of interacting networks, where each individual network aims to minimize its own error on the task but simultaneously optimize the overlap of its predictions with those of other networks in the ensemble. We explore different coupling regimes between networks during learning and find that the optimal ensemble performance occurs when the local interactions between the ensemble members drive them to be different from each other, without the need for a global goal for the ensemble. 

\section{Ensembles of learning interacting networks}

\begin{figure} [h]
  \centering
  \includegraphics[width=\textwidth]{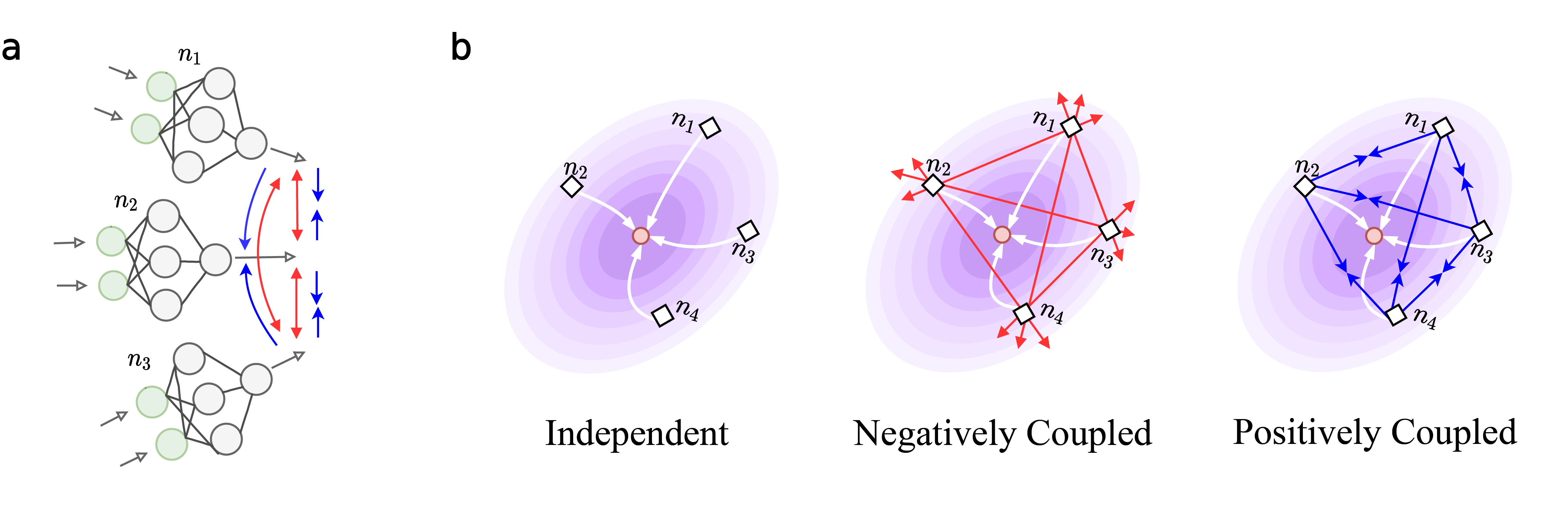}
  \caption{\textbf{Co-learning ensemble models under different coupling regimes.} \textbf{a.} An illustration of the co-learning architecture of interacting networks. The blue and red arrows designate positive or negative couplings between networks \textbf{b.} Illustrations of the three different coupling regimes between co-learning networks: (Left) For $\beta = 0$, the networks learn independently to minimize their cross-entropy loss function. Graphically, learning drives the networks (white squares) to go down the loss surface (purple ellipsoids) towards to the best performing solution (white arrows going towards the optimal solution marked by the red dot). (Middle) For $\beta < 0$, negatively coupled networks also aim to maximize their functional distance from the other networks in the ensemble, and so in addition to their drive towards to optimal solution, a `repelling force' acts between networks (red arrows). (Right) For $\beta > 0$, positively coupled networks go towards the optimal solution but an `attracting' force (blue arrows) pulls them to be functionally similar to one another.}
  \label{fig:modelcartoon}
\end{figure}

To explore the effect of interactions between networks during learning on their individual performance and that of the ensemble, we trained collections of neural networks to solve a multi-class classification task, where each network optimizes a loss function that combines its own performance and its overlap with the other networks in the ensemble. We denote the individual networks in the ensemble by $n_i$ (with $i=1...N$) and the outputs given by $n_{i}$ as $p_i(y|x)$, namely, the probability over labels $y$ for a sample $x$. We trained each of the individual networks to minimize the loss function over all samples, $\langle \mathcal{L}_i(x) \rangle_{x}$ (where $\langle \rangle_x$ denotes an average over all samples), and the loss for each sample $x$ is 
\begin{equation}
\mathcal{L}_i(x) = D_{KL}[q(y|x)||p_i(y|x)] + \sum^{N}_{j \neq i}\beta_{ij} D_{KL}[p_j(y|x)||p_i(y|x)] 
\label{eq:Li}
\end{equation}
\noindent where $q(y|x)$ is the desired classification values of $x$, $D_{KL}$ is the Kullback-Leibler divergence \cite{cover_2005} and $\beta_{ij}$ is the coupling coefficient between networks, $n_i$ and  $n_j$. Notably, for $\beta_{ij} > 0$, the loss function penalizes high $D_{KL}$ between networks, thus driving the networks to learn to give the same labels for the same input. For $\beta_{ij} < 0$, the loss function penalizes low $D_{KL}$ between networks, and its minimization implies higher diversity between networks. On each training step, the predictions of all the networks were obtained simultaneously. We note that both the forward and backward passes of each network can be done in parallel (and thus can be practically distributed over many machines). The reference point for these two coupling regimes is the case where $\beta_{ij}=0$, for which $\mathcal{L}_i(x)$ reduces to the cross entropy loss between each network and the desired label for each sample. This is akin to a collection of networks that are trained independently (using the same training set). Figure \ref{fig:modelcartoon} shows an illustration of the model. 

In the following experiments we used a uniform coupling coefficient value and all to all connectivity for the ensembles we trained. The loss function from Eq.\,\ref{eq:Li} can be re-expressed in this case as

\begin{equation} 
\mathcal{L}_i(x) = - \sum_y \left( q(y|x) \log p_i(y|x) + \beta \sum_{j \neq i}^N p_j(y|x) \log p_i(y|x)  \right) \\
 - \left( H\left[q(y|x)\right]+ \beta \sum^N_{j \neq i} H\left[p_j(y|x)\right]\right) 
\label{eq:Li_fixed_beta2}
\end{equation}

\noindent where $H\left[\cdot\right]$ is the Shannon entropy. 
Since $q(y|x)$ does not change with training, minimizing $\mathcal{L}_i(x)$ is equivalent to minimizing 
\begin{equation} 
\hat{l}_i(x) =- \sum_y\left( q(y|x) \log p_i(y|x) + \beta \sum_{j \neq i}^N p_j(y|x) \log p_i(y|x)  \right) - \beta \sum_{j \neq i} H\left[p_j(y|x)\right]
\end{equation}

\noindent After training, the ensemble's output was given by the average of the outputs of the individual networks $p_i(y|x)$, 
\begin{equation}
p_{ens}(y|x) = \frac{1}{N} \sum_{i=1}^N p_i(y|x)
\end{equation} 
\noindent The performance or accuracy of each of the individual networks and of the ensemble was evaluated on held-out test data, and quantified by the fraction of samples for which the network or the ensemble gave the highest probability to the correct label. We considered other combination schemes such as the geometric mean or majority voting, which gave similar qualitative results (see Supplementary Materials). 

\section{Results}
We used stochastic gradient descent to train collections of LeNet-5 architecture \cite{lecun_1998} using the loss function in Eq. \ref{eq:Li} on the CIFAR-10 dataset \cite{krizhevsky_2009} for 150 epochs, where networks' performance started to converge. We also studied co-learning of coupled VGG networks \cite{simonyan_2015} using CIFAR-100 dataset \cite{krizhevsky_2009} (see Supplementary Materials for implementation details). We evaluated the performance of individual networks and the ensemble for $\beta$ values in the range of $[-1, 1]$ over 30 random initialization and random dataset splits of 50,000 training samples and 10,000 test samples. We measured the dissimilarity between network $n_i$ and $n_j$ as 
\begin{equation} 
d(n_i,n_j) = \langle D_{JS}[p_i(y|x)||p_j(y|x)] \rangle_{x}
\label{eq:DJS_networks}
\end{equation}
\noindent where $D_{JS}$ is the Jensen-Shannon divergence (a bounded and symmetric similarity measure of distributions \cite{lin_1991}), and $\langle \rangle_x$ is the average over all samples in the test set. 

\begin{figure} [h]
    \centering
    \includegraphics[width=\textwidth]{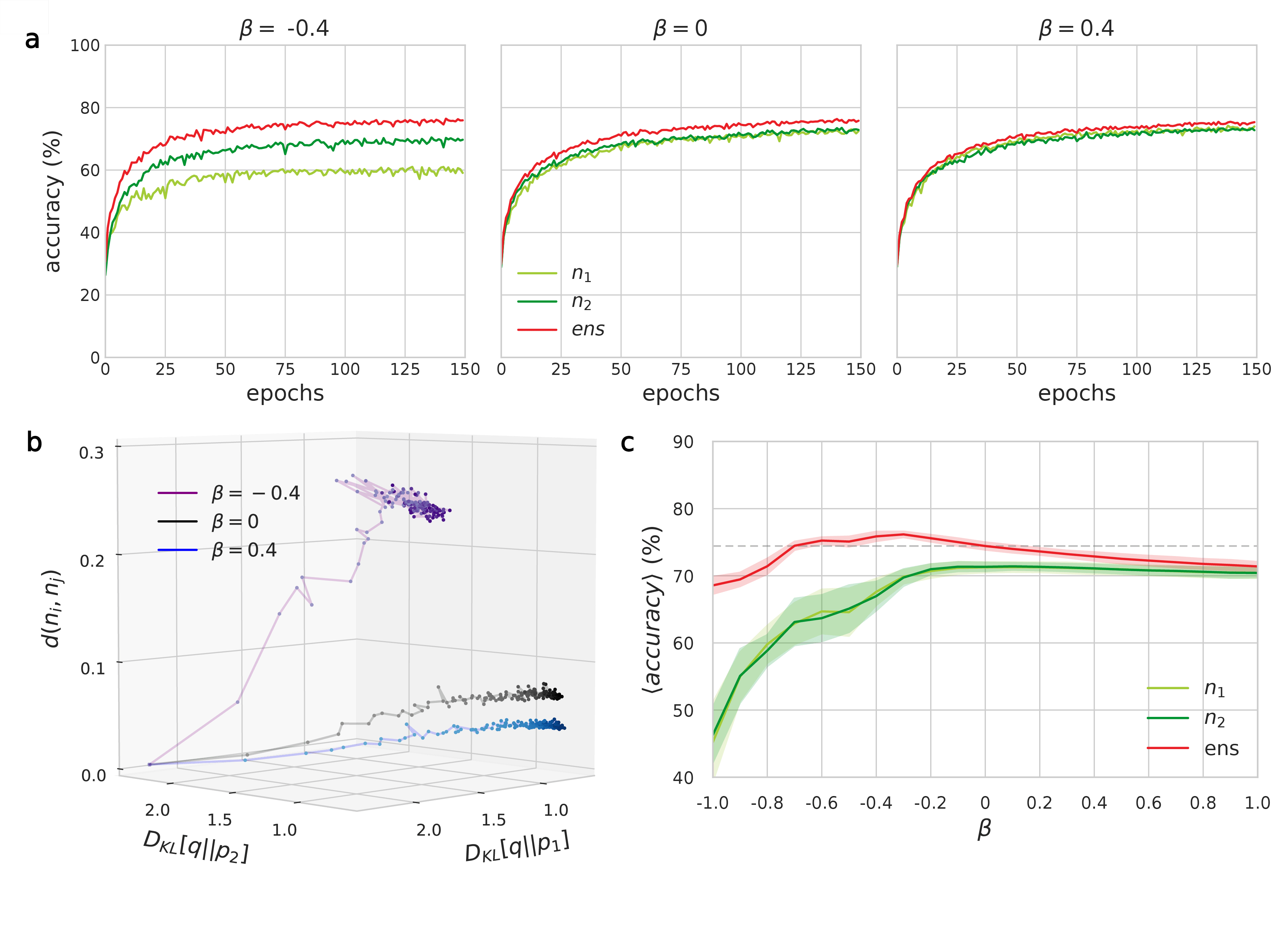}
    \caption{{\bf Learning dynamics and convergent performance levels of individual networks and small ensembles under different co-learning coupling regimes.} {\bf a.} Example of the learning curves of an ensemble of size $N=2$ and its individual networks (LeNet-5 trained on CIFAR-10); Left panel shows the case of negatively coupled networks; Middle panel shows independently trained networks, and right panel shows positively coupled ones. 
    {\bf b.} Example of the training trajectories of two networks and the distance between them is shown for ensembles from the three coupling regimes. Each sequence of points of the same color shows the values of the functional distance between networks  $d(n_i,n_j)$ against their individual performance at that point, given by $D_{KL}[q|p_i]$. Light points show the beginning of training to dark points the end of training. {\bf c.} Accuracy of two individual networks and the ensemble is shown as a function of the coupling coefficient at the end of a 150 epoch training. Lines show the average values over 30 runs with different train-test splits and random initialization; Funnels around each line show the standard deviation values over the 30 runs. Dashed horizontal line marks the performance of $\beta = 0$.  
    }
    \label{fig:2nets}
\end{figure}

Figure \ref{fig:2nets}a shows a typical example of the learning curves for one ensemble of two networks: The performance of the individual networks reaches similar accuracy for the case of independent networks, and are more similar to one another for the positively coupled case (by construction). As expected, the performance of the individual networks in the negatively coupled case decreases monotonically with more negative $\beta$ values, as reflected by their individual accuracy, $D_{KL}[q(y|x)||p_i(y|x)]$, shown in Figure \ref{fig:2nets}b. Additionally, their performances fluctuate during the training, much more than the independent or positively coupled networks.

Figure \ref{fig:2nets}c compares the performance of ensembles and the individual networks for the whole range of $\beta$ values. As is well known for the case of merging independent networks, the ensemble is more accurate than each of the individual networks. For positive coupling values the ensemble's performance decreases, and becomes more similar to each of the individual networks. The performance of the networks that were coupled decreases monotonically with more negative coupling values. Critically, the performance of the ensemble increases with negative $\beta$ values up to an optimal point, beyond which (i.e. for even more negative couplings) the ensemble performance also decreases. Thus, the most accurate ensembles are ones of negatively coupled networks. We found similar behavior for ensembles of VGG networks (see Supplementary Materials). Interestingly, these results are reminiscent of the collective behavior of the socialtaxis algorithm \cite{karpas_2017} for foraging agents: optimizing individual networks performance regularized by diversity, or increasing the diversity regularized by the task performance - give the optimal ensemble performance.

\subsection{Scaling of optimal coupling coefficient with the ensemble size}
\begin{figure} [h]
  \centering
  \includegraphics[width=\textwidth]{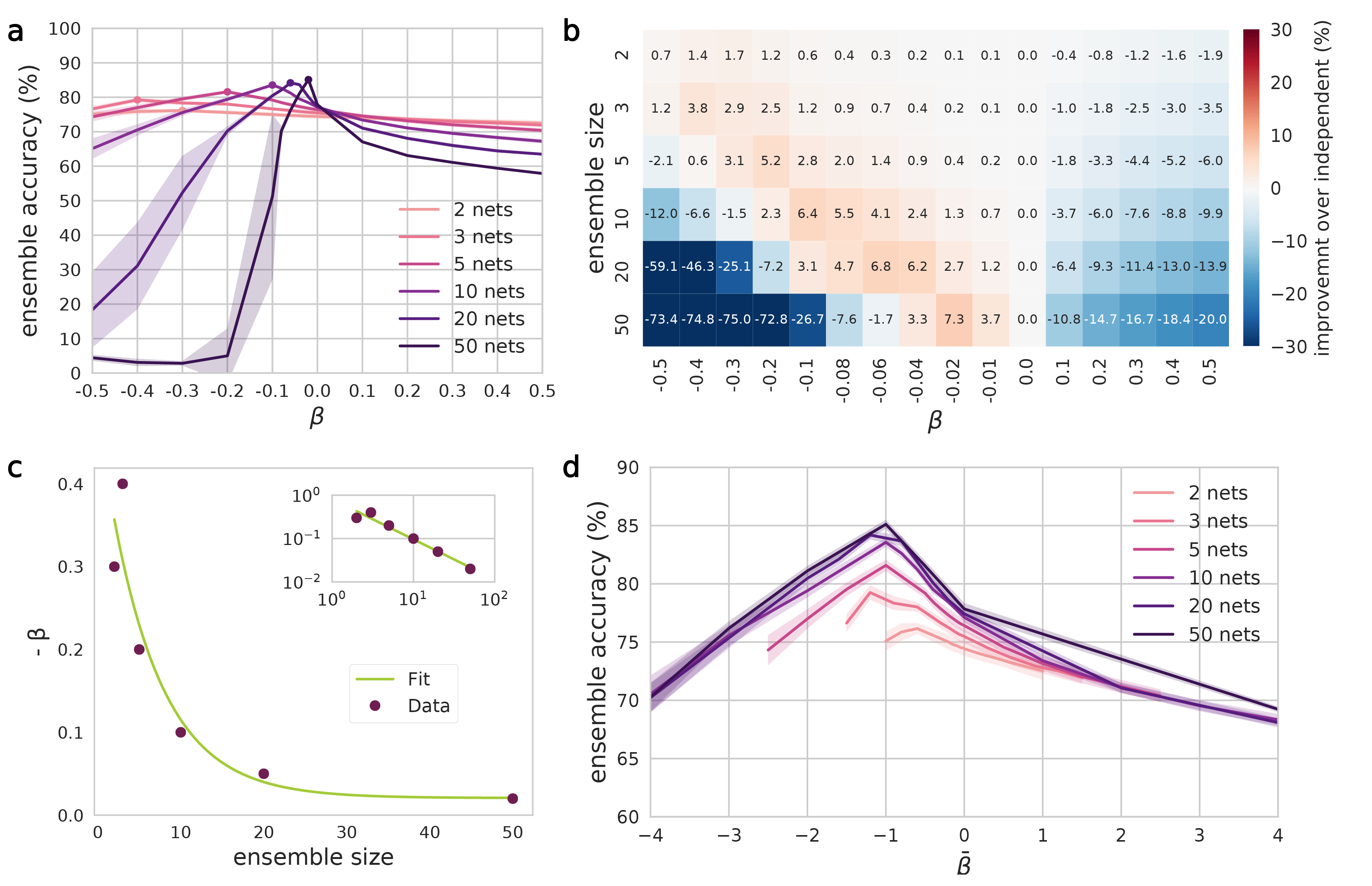}
  \caption{{\bf The optimal negative coupling of co-learning networks scales with the size of the ensemble.} {\bf a.} Average accuracy of ensembles is shown as a function of the coupling coefficient for different ensemble sizes; Funnels around the lines show the standard deviation of accuracy values over the 30 runs with different train-test splits and random initialization. ($^*$ For $N=50$ the results here were obtained from 10 repeats only). {\bf b.} Heatmap of the average difference in accuracy between coupled networks ensembles and independent ones {\bf c.} The optimal coupling $\beta$ is shown as a function of the ensemble size, with an exponential fit. Inset: optimal $\beta$ vs.\,$N$ are shown on logarithmic scale, and fit $\log(\beta) = -0.99\cdot\log(N) - 0.15$. {\bf d.} Re-plotting of the curves from panel a, but here the performance is plotted against normalized coupling values, $\bar{\beta} = \beta \cdot N$; see text.
  }
  \label{fig:largeN}
\end{figure}

Coupling more networks during training gave more accurate results for negatively coupled ensembles. Figure \ref{fig:largeN}a shows the ensemble performance against the coupling coefficient for ensembles of different sizes, $N$. The optimal $\beta$ is negative for all ensemble sizes we tested and its value depends on $N$, as reflected in Figure \ref{fig:largeN}b. While the optimally negatively coupled $N=2$ ensembles were more accurate than ensembles of independently trained networks by $1.7\%\pm0.5$, for ensembles of 50 negatively coupled networks, the optimal coupling was on average $7.3\%\pm0.2$ more accurate. The scaling of the optimal coupling coefficient with the size of the ensemble was fitted well by an exponential (Figure \ref{fig:largeN}c); on logarithmic scale it is easy to extract a slope of $-0.99\pm0.18$, reflecting that $\beta$ scales nearly linearly with $\frac{1}{N}$. We thus replace $\beta$ in equation \ref{eq:Li} with $\frac{\bar{\beta}}{N}$ and re-plot the performance of the ensembles for different values of $N$ as a function $\bar{\beta}$, and find that the peaks for different values of $N$ is around $\bar{\beta} \approx -1$ (Figure \ref{fig:largeN}d).

\subsection{Functional dissimilarity of negatively coupled networks}
\begin{figure} [h]
  \centering
  \includegraphics[width=\textwidth]{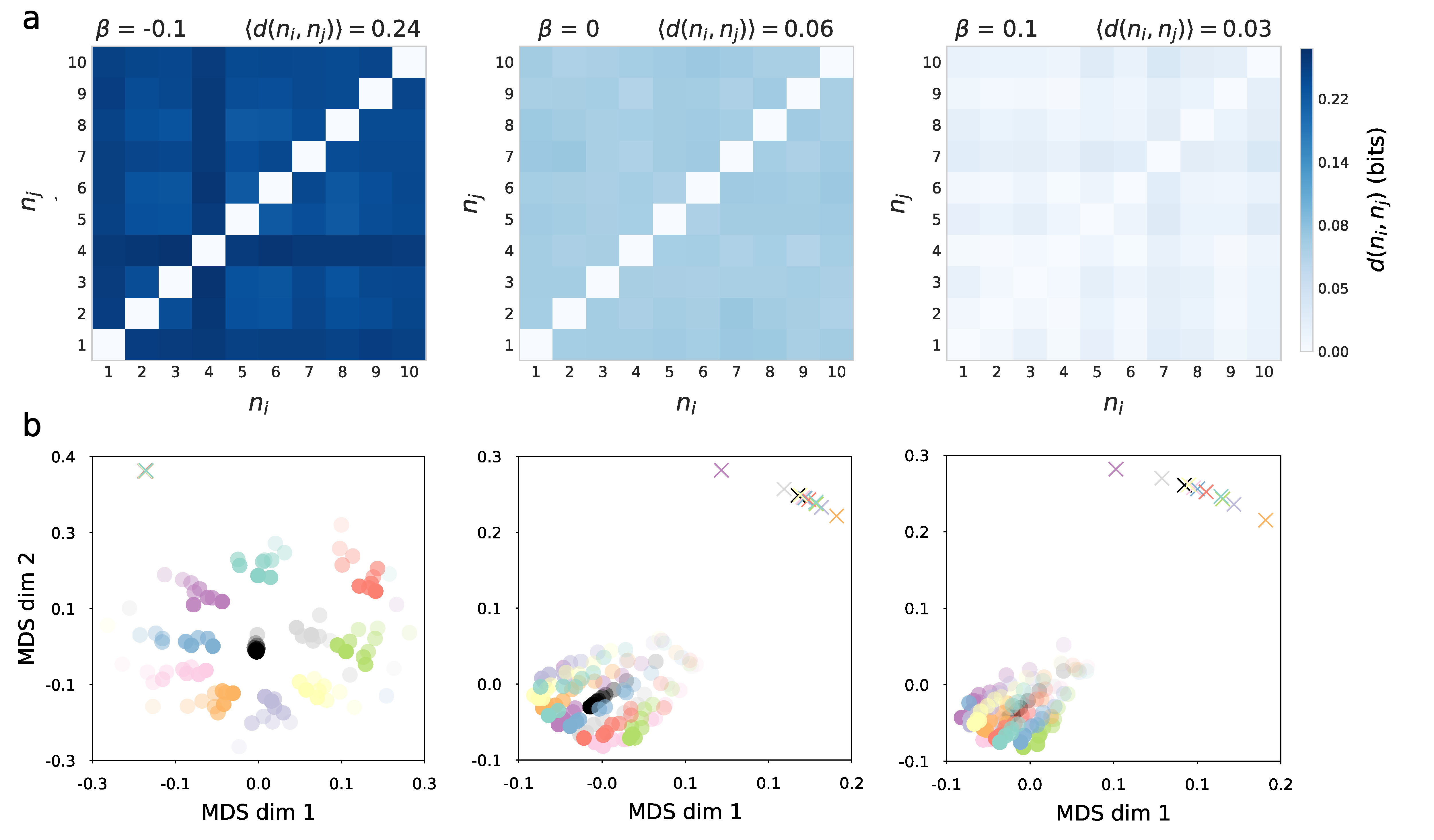}
  \caption{{\bf Dynamics of individual networks in functional space during learning shows diverging structure and organization of negatively coupled co-learning networks.} {\bf a.} Each matrix shows the functional dissimilarity $d(n_i,n_j)$ between all pairs of networks in an ensemble of size $N$ trained with different couplings at the end of training of one run. Similar results were obtained on different repeats. Ensembles trained with $\beta < 0$ have much higher pairwise functional distances than independently trained networks and positively coupled ensembles. {\bf b.} Multidimensional scaling embedding of the networks and the ensemble, based on the functional dissimilarity between networks, along a single training trajectory. Each color denotes one particular network during training, and black dot marks the ensemble. Crosses indicate the starting point of each network or the ensemble. Color intensity shows the progression with training (low to high). Correlations between $d(n_i,n_j)$ and the distances between the networks in the MDS embedding were: $r=0.78, p<0.005$; $r=0.95, p<0.005$; $r=0.98, p<0.005$, for negatively coupled, independent and positively coupled ensembles respectively.}
  \label{fig:MDS}
\end{figure}

The networks that make the ensembles trained using negative coupling, converge at the end of training to give diverse predictions, by construction. Figure \ref{fig:MDS}a shows the pairwise functional dissimilarity, $d(n_i,n_j)$ at the end of training between the networks that make an ensemble of size $N=10$, for three examples of couplings. Networks of negatively coupled ensembles show much higher pairwise distances compared to the ones in independent and positively coupled ensembles. Figure \ref{fig:MDS}b shows the dynamics of the functional similarity between networks and the ensemble during training, by embedding the networks based on the functional distances $d(n_i,n_j)$ between them using Multi-Dimensional Scaling (MDS) \cite{kruskal_1978}. Negatively coupled networks separate (repel) early on and then converge to a layout where they `surround' $p_{ens}$. Positively coupled networks and the independent are tightly packed around $p_{ens}$ and the trajectories of individual networks overlap with that of the ensemble.

\begin{figure} [h]
  \centering
  \includegraphics[width=\textwidth]{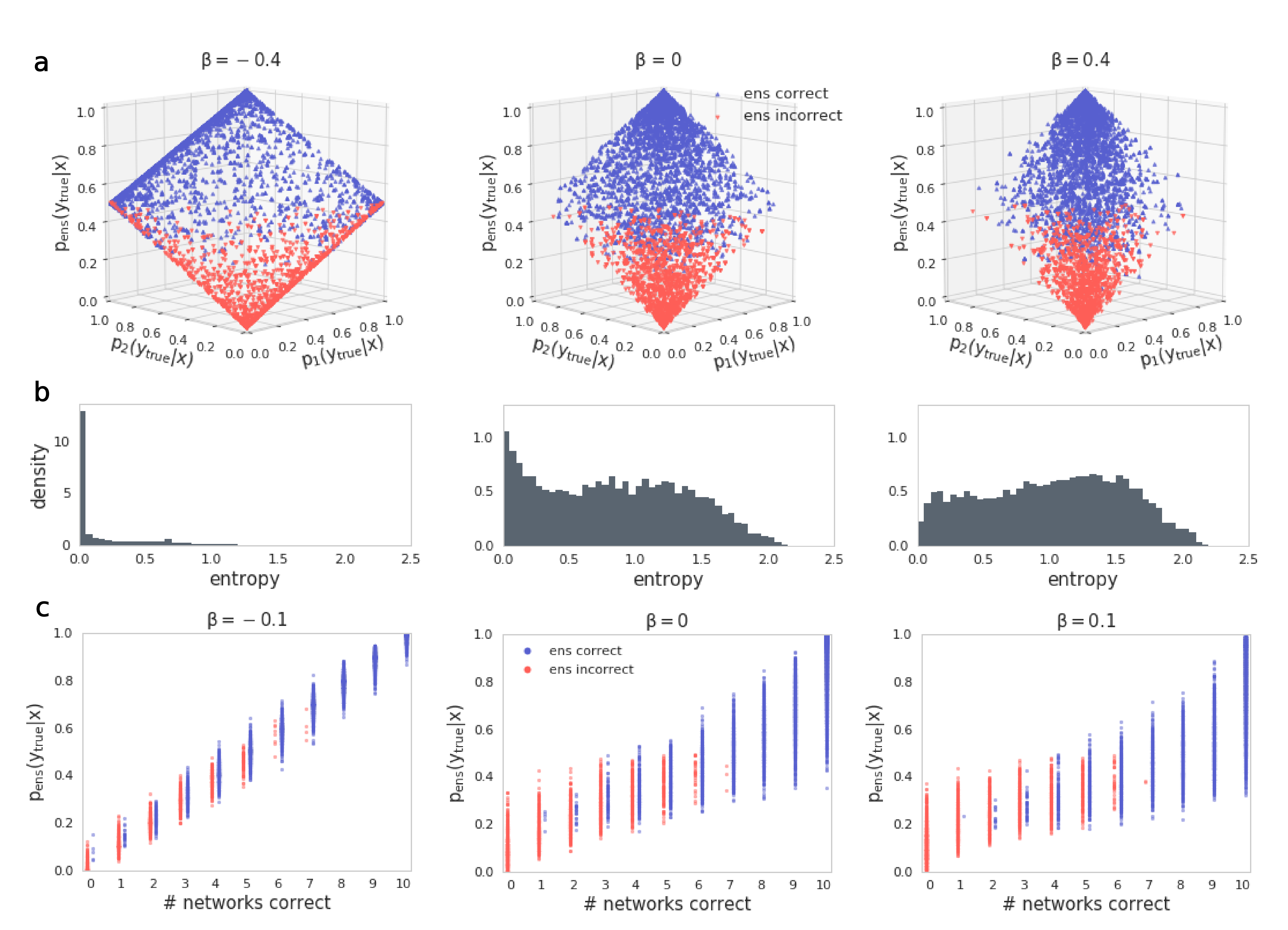}
  \caption{{\bf Dissimilarity of networks on specific samples elucidates specialization of networks in negatively coupled ensembles.} {\bf a.} Panels show the performance of networks over individual samples for different $\beta$ value of ensembles of size $N=2$. Each dot shows the classification values given to each sample by individual networks, $p_1(y_{true}|x)$, $p_2(y_{true}|x)$, and the ensemble, $p_{ens}(y_{true}|x)$. Dots' color designates correct (blue) or incorrect (red) prediction of the ensemble. Notably, $\beta<0$ coupling drives the dots towards the `walls' of the cube, where one network predicts the classification value with high probability and the other one doesn't. 
  {\bf b.} Representative examples of the histograms of entropy values of the classifications of samples by one network in the ensemble, $H[p_i(y|x)]$, for $N=2$. Negatively coupled networks have high `confidence' in their assessments, reflected by the large fraction of samples with low entropy. {\bf c.} For ensembles of size $N=10$, the prediction of the ensemble on each sample is shown vs the number of networks in the ensemble that would classify it correctly. Dots' color indicates whether the ensemble would classify this sample correctly (blue) or not (red). (Red and blue dots were shifted slightly sideways to show the range of values of sample classifications). For all $\beta$ values there is a strong correlation between the number of networks that give the correct label and the ensemble's confidence, with significantly tighter correlation for $\beta<0$.}
  \label{fig:sampledivergence}
\end{figure}

We further explored the similarity of networks in the ensemble in terms of their overlap over specific samples. Figure \ref{fig:sampledivergence}a shows the agreement over samples among the individual networks and the ensemble for $N=2$, under different coupling regimes. We found that for negatively coupled networks, more samples had high likelihood value for one of the networks and not the other, whereas the likelihood value of the ensemble was high enough to make the correct prediction (reflected by the fraction of samples at the edges on panel a). We measured the `confidence' or certainty of the networks' classification in each case by the distribution of entropy values of one of the networks over the samples, $H[p_i(y|x)]$. Figure \ref{fig:sampledivergence}b shows that negatively coupled networks had a sharp peak close to 0, and the range of entropy values was larger for independent networks or $\beta>0$ (For similar result for $N=10$ see Supplementary Materials). Thus, negatively coupled networks became more confident in their classification of samples. 

The difference in classification values or the confidence of individual networks is also manifested in the distribution of `opinions' over the networks in the ensembles for the same sample. Figure \ref{fig:sampledivergence}c shows the ensemble's likelihood of giving the correct answer against the number of individual networks ($N=10$) that voted for the correct answer. We found that negative coupling creates a tight correlation ($r_s = 0.99, p<0.005$) between the ensemble's confidence in the right classification and the number of networks in the ensemble that share the same opinion. Lower correlation observed in independent ($r_s = 0.92, p<0.005$) and positively coupled ensembles ($r_s = 0.90, p<0.005$). 
Interestingly, in negatively coupled ensembles, there are some data samples on which none of the individual networks was correct and yet the ensemble was right.

\subsection{Diversity of network parameters in ensembles of coupled networks}
\begin{figure} [h]
  \centering
  \includegraphics[width=\textwidth]{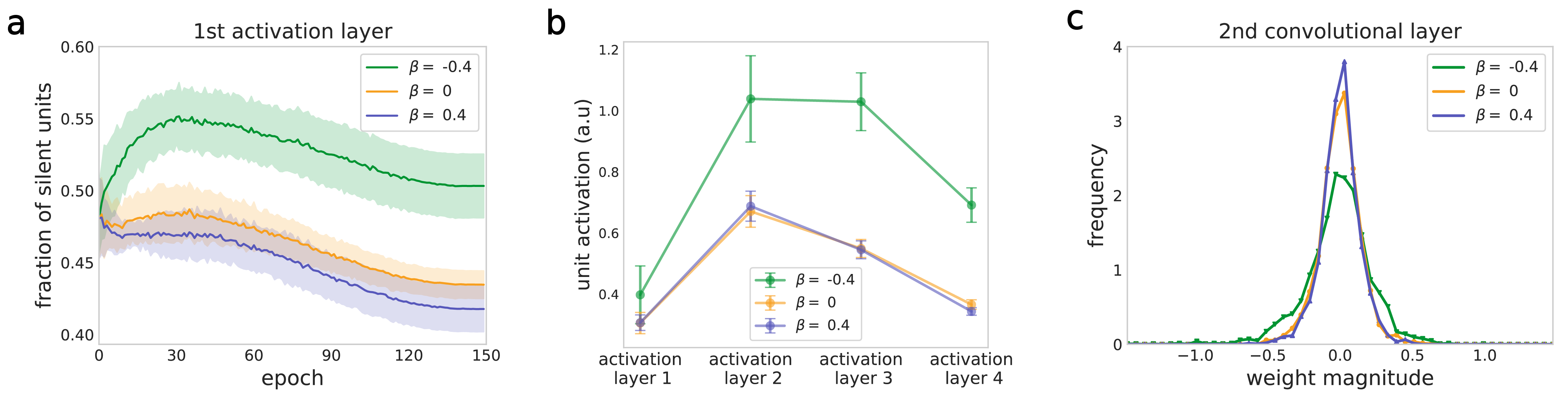}
  \caption{{\bf Networks in negatively co-learning ensembles show structural differences in their connectivity patterns and sparseness of activation of neurons.} \textbf{a.} Sparseness of activation of neurons in the first hidden layer of one exemplary network is shown during training by the mean fraction of non-activated units per sample. Negatively coupled networks show sparser activity during learning and in their final configuration. \textbf{b.} The mean activation per sample for neuron at each of the hidden layers (averaged over 30 train-test splits and random initialization). Negatively coupled networks reach higher firing rates per sample \textbf{c.}. A typical example of the histograms of weight values from the second convolutional layer in one network, showing that negatively coupled networks utilized wider weight distributions compared to independent and positively coupled ones.}
  \label{fig:hidden-activations}
\end{figure}

We asked what are the effects of different coupling among the individual networks in the ensemble during learning and on their parameters. First, we measured the sparseness of activated units in each hidden layer in the network over all samples in the test sets. Figure \ref{fig:hidden-activations}a shows the fraction of non-activated units in the first hidden layer for different coupling regimes, during the training for networks in ensembles of size $N=2$. Negatively coupled networks had higher fraction of inactive units, whereas positively coupled networks had lower fraction of inactive neurons compared to independent networks. Negatively coupled networks had higher mean activation in all hidden layers (Figure \ref{fig:hidden-activations}b). Furthermore, Figure \ref{fig:hidden-activations}c shows that negatively coupled networks utilized wider distribution of weights in all layers of the networks (similar results obtained in other layers, see Supplementary Materials). 
\clearpage

\subsection{Expanding and boosting ensembles of co-learning networks}
Finally, we asked how adding new networks to an ensemble of co-learning networks could change its performance. As an example, we compared the performance of ensembles of 10 networks trained with their optimal negative $\beta$, to ensembles that combined these 10 networks with new 5 networks in different ways. We found that adding 5 independently trained networks or adding 5 networks that are negatively coupled to the first 10 networks did not improve the ensemble's performance. Simple boosting of the ensemble by training the new 5 networks on samples that were close to the decision boundary of the original ensemble did not improve the ensemble's performance either. In all cases, retraining 15 networks by negatively coupling them was a significantly better strategy (see Supplementary Materials). We conclude that iterative expansions of coupled networks or boosting requires more delicate tuning or learning. 

\section{Discussion}
Inspired by analysis and models of social interactions in groups of animals, we presented a model for co-learning in ensembles of interacting deep neural networks. We found that for a wide range of ensemble sizes, the performance of the ensemble was most accurate when individual networks aimed to maximize their individual performance as well as their diversity. In particular, optimal ensemble co-learning by coupled networks resulted in individual performance of networks that was significantly inferior to what the typical network would achieve if it was trained on its own. We showed that ensembles of diverging networks reach their optimal behavior by specialization of individual networks on parts of the space of samples, and structural features that are distinct from that of independently trained or positively coupled co-learning networks.  

The models for co-learning that we studied here reflect the importance of diversity as a design principle for ensemble learning or efficient group behavior, and the potential benefits of optimizing this diversity. We relied on uniform interaction coefficient for all pairs of networks in the ensemble, which implies that adapting or learning the `social' coefficients could further improve the performance of the ensemble. The specialization of individual networks in optimally interacting ensembles that we observed, suggests possible boosting of ensembles (beyond our simple attempt described above), by adding networks with different interaction terms or targeted samples. Future work will also explore larger ensembles, other network architectures, and different task difficulties. Finally, similar to the extension of game theory ideas to the use of generative adversarial networks \cite{goodfellow_2014}, we hypothesize that balancing conflicts between learning networks and diversification, as well as considering settings with partial and missing information among networks, could lead to better training and performance of ensembles. 

\small

\medskip

\section*{Acknowledgments}
We thank Udi Karpas, Tal Tamir, Yoni Mayzel, Omri Camus, Adam Haber, Or Samimi Golan, Or Ram, and Allan Drummond for discussions, comments, and ideas.
This work was supported by the Simons Collaboration on the Global Brain (542997), and the Israel–US Binational Science Foundation. Additional support for this work came from Martin Kushner Schnur and Mr. and Mrs. Lawrence Feis; ES is the Joseph and Bessie Feinberg Professorial Chair.

\bibliography{references4arxiv}

\begin{thebibliography}{23}
\providecommand{\natexlab}[1]{#1}
\providecommand{\url}[1]{\texttt{#1}}
\expandafter\ifx\csname urlstyle\endcsname\relax
  \providecommand{\doi}[1]{doi: #1}\else
  \providecommand{\doi}{doi: \begingroup \urlstyle{rm}\Url}\fi

\bibitem[Breiman(1996)]{breiman_1996}
Leo Breiman.
\newblock Bagging predictors.
\newblock \emph{Machine learning}, 24\penalty0 (2):\penalty0 123--140, aug
  1996.
\newblock ISSN 0885-6125.
\newblock \doi{10.1007/{BF00058655}}.
\newblock URL \url{http://link.springer.com/10.1007/{BF00058655}}.

\bibitem[Freund and Schapire(1996)]{freund_1996}
Yoav Freund and Robert~E. Schapire.
\newblock Experiments with a new boosting algorithm.
\newblock \emph{In Proceedings of the 13th International Conference on Machine
  Learning}, 96:\penalty0 148--156, 1996.
\newblock URL
  \url{https://citeseerx.ist.psu.edu/viewdoc/summary?doi=10.1.1.297.5231}.

\bibitem[Webb et~al.(2019)Webb, Reynolds, Iliescu, Reeve, Luján, and
  Brown]{webb_2019}
Andrew~Michael Webb, Charles Reynolds, Dan-Andrei Iliescu, Henry Reeve, Mikel
  Luján, and Gavin Brown.
\newblock Joint training of neural network ensembles.
\newblock \emph{Unpublished}, 2019.
\newblock \doi{10.13140/rg.2.2.28091.46880}.
\newblock URL \url{http://rgdoi.net/10.13140/{RG}.2.2.28091.46880}.

\bibitem[Izmailov et~al.(2018)Izmailov, Podoprikhin, Garipov, Vetrov, and
  Wilson]{izmailov_2018}
Pavel Izmailov, Dmitrii Podoprikhin, Timur Garipov, Dmitry Vetrov, and
  Andrew~Gordon Wilson.
\newblock Averaging weights leads to wider optima and better generalization.
\newblock \emph{Name 34th Conference on Uncertainty in Artificial
  Intelligence}, jan 2018.
\newblock URL
  \url{https://nyuscholars.nyu.edu/en/publications/averaging-weights-leads-to-wider-optima-and-better-generalization}.

\bibitem[Huang et~al.(2017)Huang, Li, Pleiss, Liu, Hopcroft, and
  Weinberger]{huang_2017}
Gao Huang, Yixuan Li, Geoff Pleiss, Zhuang Liu, John~E. Hopcroft, and Kilian~Q.
  Weinberger.
\newblock Snapshot ensembles: Train 1, get m for free.
\newblock \emph{5th International Conference on Learning Representations},
  2017.

\bibitem[Garipov et~al.(2018)Garipov, Izmailov, Podoprikhin, Vetrov, and
  Wilson]{garipov_2018}
Timur Garipov, Pavel Izmailov, Dmitrii Podoprikhin, Dmitry Vetrov, and
  Andrew~Gordon Wilson.
\newblock Loss surfaces, mode connectivity, and fast ensembling of {DNNs}.
\newblock \emph{Advances in Neural Information Processing Systems}, 2018.
\newblock URL \url{https://github.com/timgaripov/dnn-mode-connectivity}.

\bibitem[Dutt et~al.(2019)Dutt, Pellerin, and Quénot]{dutt_2019}
Anuvabh Dutt, Denis Pellerin, and Georges Quénot.
\newblock Coupled ensembles of neural networks.
\newblock \emph{Neurocomputing}, apr 2019.
\newblock ISSN 09252312.
\newblock \doi{10.1016/j.neucom.2018.10.092}.
\newblock URL
  \url{https://linkinghub.elsevier.com/retrieve/pii/S0925231219304370}.

\bibitem[Ramdya et~al.(2015)Ramdya, Lichocki, Cruchet, Frisch, Tse, Floreano,
  and Benton]{ramdya_2015}
Pavan Ramdya, Pawel Lichocki, Steeve Cruchet, Lukas Frisch, Winnie Tse, Dario
  Floreano, and Richard Benton.
\newblock Mechanosensory interactions drive collective behaviour in drosophila.
\newblock \emph{Nature}, 519\penalty0 (7542):\penalty0 233--236, mar 2015.
\newblock \doi{10.1038/nature14024}.
\newblock URL \url{http://dx.doi.org/10.1038/nature14024}.

\bibitem[Strandburg-Peshkin et~al.(2015)Strandburg-Peshkin, Farine, Couzin, and
  Crofoot]{strandburgpeshkin_2015}
Ariana Strandburg-Peshkin, Damien~R Farine, Iain~D Couzin, and Margaret~C
  Crofoot.
\newblock Shared decision-making drives collective movement in wild baboons.
\newblock \emph{Science}, 348\penalty0 (6241):\penalty0 1358--1361, jun 2015.
\newblock \doi{10.1126/science.aaa5099}.
\newblock URL \url{http://dx.doi.org/10.1126/science.aaa5099}.

\bibitem[Gelblum et~al.(2015)Gelblum, Pinkoviezky, Fonio, Ghosh, Gov, and
  Feinerman]{gelblum_2015}
Aviram Gelblum, Itai Pinkoviezky, Ehud Fonio, Abhijit Ghosh, Nir Gov, and Ofer
  Feinerman.
\newblock Ant groups optimally amplify the effect of transiently informed
  individuals.
\newblock \emph{Nature Communications}, 6:\penalty0 7729, jul 2015.
\newblock \doi{10.1038/ncomms8729}.
\newblock URL \url{http://dx.doi.org/10.1038/ncomms8729}.

\bibitem[Karpas et~al.(2017)Karpas, Shklarsh, and Schneidman]{karpas_2017}
{ED} Karpas, A~Shklarsh, and E~Schneidman.
\newblock Information socialtaxis and efficient collective behavior emerging in
  groups of information-seeking agents.
\newblock \emph{Proceedings of the National Academy of Sciences of the United
  States of America}, 114\penalty0 (22):\penalty0 5589--5594, may 2017.
\newblock \doi{10.1073/pnas.1618055114}.
\newblock URL \url{http://dx.doi.org/10.1073/pnas.1618055114}.

\bibitem[Livnat and Pippenger(2006)]{livnat_2006}
A~Livnat and N~Pippenger.
\newblock An optimal brain can be composed of conflicting agents.
\newblock \emph{Proceedings of the National Academy of Sciences of the United
  States of America}, 103\penalty0 (9):\penalty0 3198--3202, feb 2006.
\newblock \doi{10.1073/pnas.0510932103}.
\newblock URL \url{http://dx.doi.org/10.1073/pnas.0510932103}.

\bibitem[Cover and Thomas(2005)]{cover_2005}
Thomas~M. Cover and Joy~A. Thomas.
\newblock \emph{Elements of Information Theory}.
\newblock John Wiley \& Sons, Inc., Hoboken, {NJ}, {USA}, sep 2005.
\newblock ISBN 9780471241959.
\newblock \doi{10.1002/{047174882X}}.
\newblock URL \url{http://doi.wiley.com/10.1002/{047174882X}}.

\bibitem[Lecun et~al.(1998)Lecun, Bottou, Bengio, and Haffner]{lecun_1998}
Y.~Lecun, L.~Bottou, Y.~Bengio, and P.~Haffner.
\newblock Gradient-based learning applied to document recognition.
\newblock \emph{Proceedings of the {IEEE}}, 86\penalty0 (11):\penalty0
  2278--2324, 1998.
\newblock ISSN 00189219.
\newblock \doi{10.1109/5.726791}.
\newblock URL
  \url{http://ieeexplore.ieee.org/lpdocs/epic03/wrapper.htm?arnumber=726791}.

\bibitem[Krizhevsky(2009)]{krizhevsky_2009}
Alex Krizhevsky.
\newblock Learning multiple layers of features from tiny images.
\newblock Technical report, 2009.
\newblock URL \url{https://www.cs.toronto.edu/\~kriz/cifar.html}.

\bibitem[Simonyan and Zisserman(2015)]{simonyan_2015}
Karen Simonyan and Andrew Zisserman.
\newblock Very deep convolutional networks for large-scale image recognition.
\newblock Technical report, 2015.
\newblock URL \url{http://www.robots.ox.ac.uk/}.

\bibitem[Lin(1991)]{lin_1991}
J.~Lin.
\newblock Divergence measures based on the shannon entropy.
\newblock \emph{{IEEE} Transactions on Information Theory}, 37\penalty0
  (1):\penalty0 145--151, 1991.
\newblock ISSN 00189448.
\newblock \doi{10.1109/18.61115}.
\newblock URL \url{http://ieeexplore.ieee.org/document/61115/}.

\bibitem[Kruskal and Wish(1978)]{kruskal_1978}
Joseph Kruskal and Myron Wish.
\newblock \emph{Multidimensional Scaling}.
\newblock {SAGE} Publications, Inc., 2455 Teller Road, Thousand
  Oaks California 91320 United States of America , 1978.
\newblock ISBN 9780803909403.
\newblock \doi{10.4135/9781412985130}.
\newblock URL \url{http://methods.sagepub.com/book/multidimensional-scaling}.

\bibitem[Goodfellow et~al.(2014)Goodfellow, Pouget-Abadie, Mirza, Xu,
  Warde-Farley, Ozair, Courville, and Bengio]{goodfellow_2014}
I~Goodfellow, J~Pouget-Abadie, M~Mirza, B~Xu, D~Warde-Farley, S~Ozair,
  A~Courville, and Y~Bengio.
\newblock Generative adversarial nets.
\newblock \emph{Advances in Neural Information Processing Systems}, page 2672,
  2014.
\newblock URL
  \url{http://papers.nips.cc/paper/5423-generative-adversarial-nets}.

\bibitem[Glorot and Bengio(2010)]{glorot_2010}
Xavier Glorot and Yoshua Bengio.
\newblock Understanding the difficulty of training deep feedforward neural
  networks.
\newblock In Yee~Whye Teh and Mike Titterington, editors, \emph{Proceedings of
  the Thirteenth International Conference on Artificial Intelligence and
  Statistics}, volume~9 of \emph{Proceedings of Machine Learning Research},
  pages 249--256, Chia Laguna Resort, Sardinia, Italy, 13--15 May 2010. PMLR.
\newblock URL \url{http://proceedings.mlr.press/v9/glorot10a.html}.

\bibitem[Loshchilov and Hutter(2017)]{loshchilov_2017}
Ilya Loshchilov and Frank Hutter.
\newblock {SGDR}: Stochastic gradient descent with warm restarts.
\newblock \emph{5th International Conference on Learning Representations},
  2017.
\newblock URL \url{https://arxiv.org/abs/1608.03983}.

\bibitem[Krizhevsky et~al.(2012)Krizhevsky, Sutskever, and
  Hinton]{krizhevsky_2012}
A~Krizhevsky, I~Sutskever, and {GE} Hinton.
\newblock Imagenet classification with deep convolutional neural networks.
\newblock \emph{Advances in Neural Information Processing Systems}, 2012.
\newblock URL
  \url{http://papers.nips.cc/paper/4824-imagenet-classification-with-deep-convolutional-neural-networ}.

\bibitem[Paszke et~al.(2019)Paszke, Gross, Massa, Lerer, Bradbury, Chanan,
  Killeen, Lin, Gimelshein, Antiga, Desmaison, Kopf, Yang, DeVito, Raison,
  Tejani, Chilamkurthy, Steiner, Fang, Bai, and Chintala]{pytorch_2019}
Adam Paszke, Sam Gross, Francisco Massa, Adam Lerer, James Bradbury, Gregory
  Chanan, Trevor Killeen, Zeming Lin, Natalia Gimelshein, Luca Antiga, Alban
  Desmaison, Andreas Kopf, Edward Yang, Zachary DeVito, Martin Raison, Alykhan
  Tejani, Sasank Chilamkurthy, Benoit Steiner, Lu~Fang, Junjie Bai, and Soumith
  Chintala.
\newblock Pytorch: An imperative style, high-performance deep learning library.
\newblock In H.~Wallach, H.~Larochelle, A.~Beygelzimer, F.~d'Alch\'{e} Buc,
  E.~Fox, and R.~Garnett, editors, \emph{Advances in Neural Information
  Processing Systems 32}, pages 8024--8035. Curran Associates, Inc., 2019.
\newblock URL
  \url{http://papers.neurips.cc/paper/9015-pytorch-an-imperative-style-high-performance-deep-learning-library.pdf}.

\end{thebibliography}

\normalsize

\newpage 

\section*{Supplementary Materials}
\beginsupplement

These supplementary materials present: (1) the implementation details and training of the ensembles of the neural networks presented in the paper
, (2)  the performance of other ensemble combination methods such as majority voting and the geometric mean of the networks outputs
, (3)  performance results with a VGG architecture trained on the CIFAR-100 dataset
, (4)  summary of differences in sparseness, activation, synaptic weights, and predictions in networks trained with different coupling regimes
, (5) preliminary results of the experiments of expanding an existing ensemble or boosting it. 

\subsection*{Neural Networks architectures and training details}
\label{Architectures}
\begin{figure} [h]
  \centering
  \includegraphics[width=\textwidth]{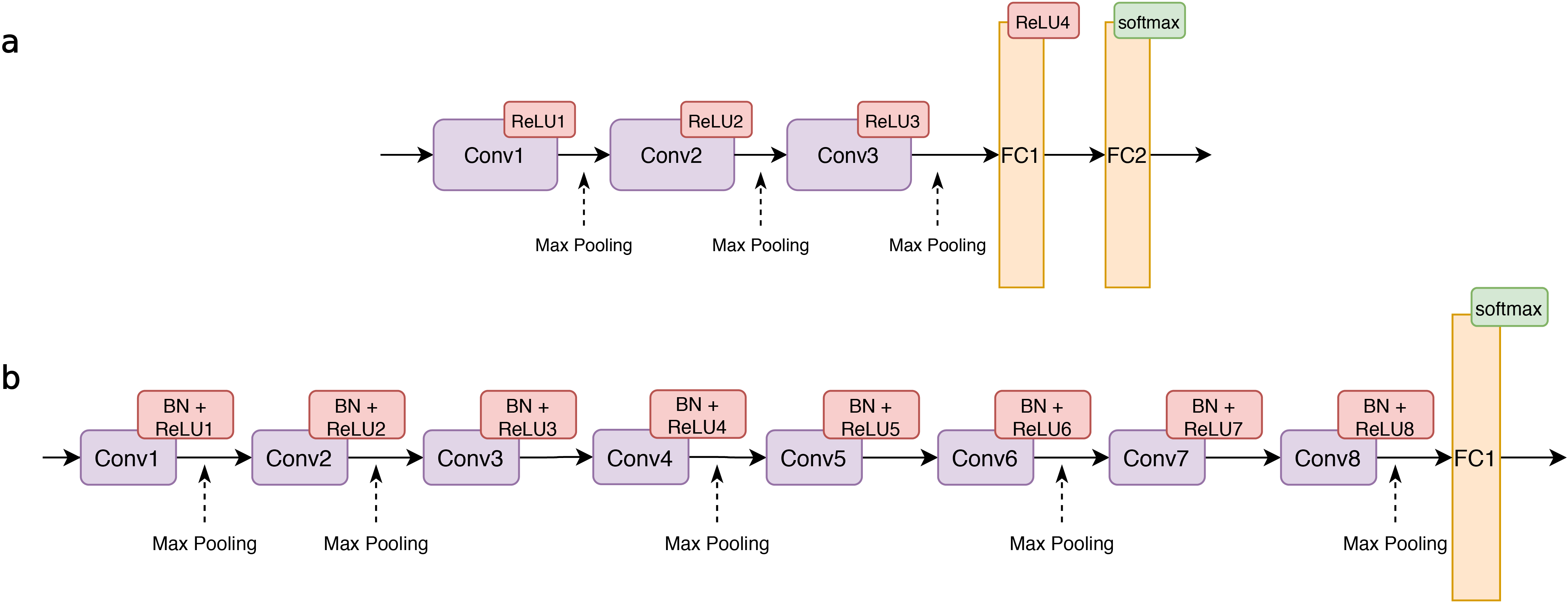}
  \caption{{\bf The architectures studied in this paper.} Convolutional layers are denoted by `Conv', fully-connected layers by `FC' and batch-normalization by `BN'. ReLU is the standard ReLU activation function and Max-Pooling is a sub-sampling process. {\bf a.} The LeNet-5 architecture, details are in Table \ref{table:lenet_architecture}. {\bf b.} The VGG-9 architecture, details are in Table \ref{table:vgg_architecture}. The architecture is similar to VGG-11 \cite{simonyan_2015}, but with only one fully-connected layer at the end of the networks instead of 3.}
  \label{fig:architecture}
\end{figure}

\begin{figure} [h]
  \centering
  \includegraphics[width=\textwidth]{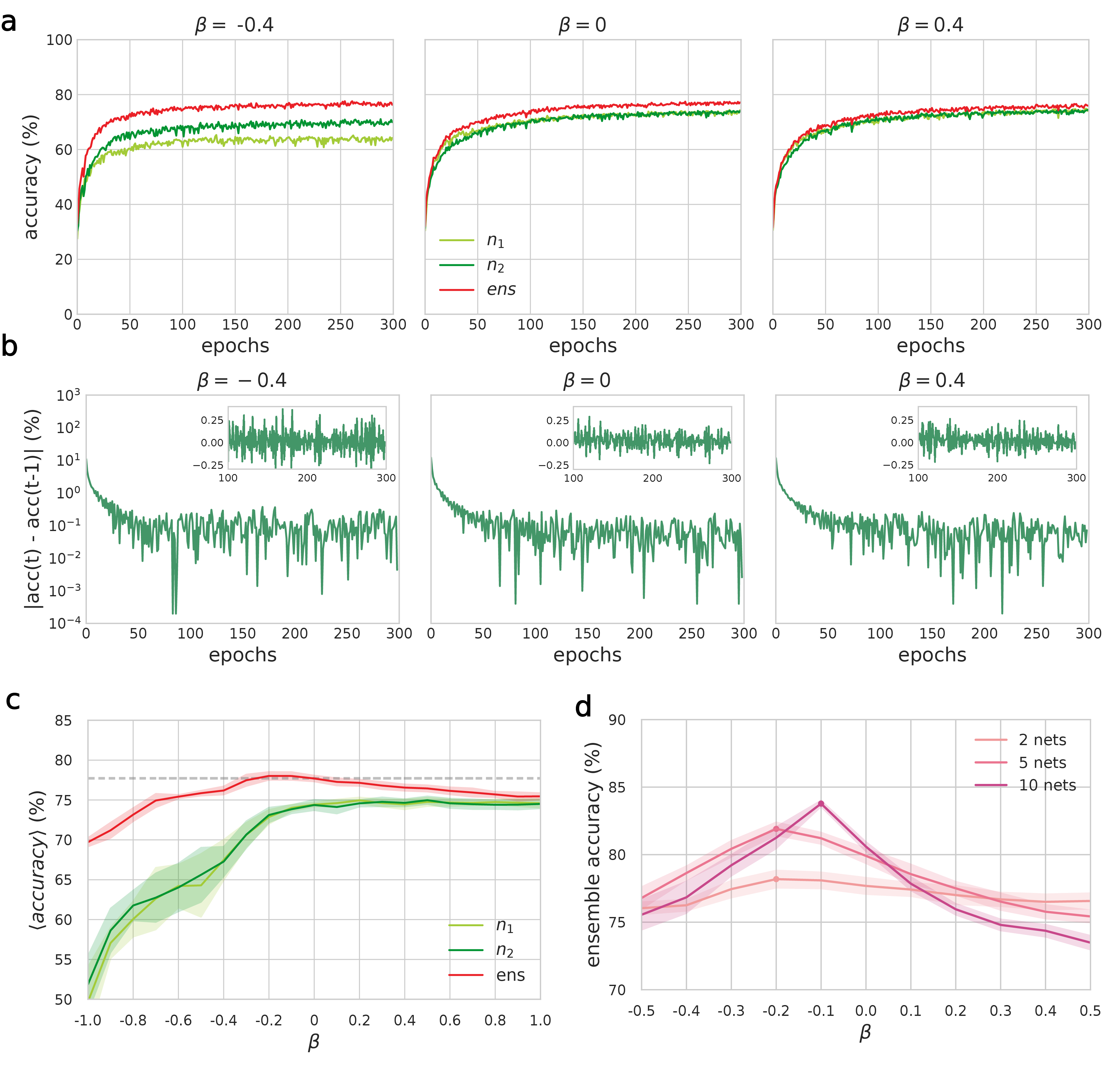}
  \caption{Negatively coupled ensembles reach higher performance compared to ensembles of independent networks and positively coupled ones for training over 300 epochs. {\bf a.} Example of the training curves of an ensemble of size $N=2$ and its individual networks (LeNet-5 trained on CIFAR-10). Left panel shows the case of negatively coupled networks; Middle panel shows independently trained networks; Right panel shows positively coupled ones. {\bf b.} The absolute change (on log-scale) in accuracy between two consecutive epochs along the training trajectory for one of the networks in an ensemble of size $N=2$; Inset shows the last 200 epochs on linear scale. {\bf c.} Accuracy of two individual networks and the ensemble is shown as a function of the coupling coefficient at the end of a 300 epoch training. Lines show the average values over 10 runs with different train-test splits and random initialization; Funnels around each line show the standard deviation values over the 10 runs. Dashed horizontal line marks the performance of $\beta = 0$. {\bf d.} Average accuracy of ensembles is shown as a function of the coupling coefficient for different ensemble sizes at the end of a 300 epoch training. Funnels around each line show the standard deviation values over the 10 runs.}
  \label{fig:300_epochs_results}
\end{figure}

Two neural network architectures and data sets were used for studying co-learning in ensembles described in the main text. 
LeNet-5 networks with ReLU activation functions (Figure \ref{fig:architecture}a and Table \ref{table:lenet_architecture}) were trained on the CIFAR-10 data set. The networks' parameters were initialized with the Xavier initialization procedure \cite{glorot_2010}. The training was performed using mini-batch stochastic gradient descent with batch size of 512, momentum of 0.9, and weight decay of 0.0005. Results presented in the main text utilized a constant learning rate of 0.01, but similar behavior was observed for different learning rate schedules such as a step function and cosine annealing: $\eta_t = \frac{1}{2}\eta_0(1+\cos{\frac{t}{t_{max}}\pi})$, where $\eta_t$ is the learning rate at epoch $t$ and $t_{max}$ is total number of epochs \cite{loshchilov_2017}. Networks were trained for 150 or 300 epochs, based on the typical convergence rate of training, as shown in Figure \ref{fig:300_epochs_results}b. The results of training for both network architectures and data sets were qualitatively similar. Negatively coupled networks achieved better performance than independent ones while the performance of their individual networks degraded with more negative coupling coefficient values (Figure \ref{fig:300_epochs_results}c). This effect was greater in larger ensembles and the optimal coupling coefficient scaled with the number of networks in the ensemble (Figure \ref{fig:300_epochs_results}d).

\begin{table}[h]
    \caption{\textbf{LeNet-5 architecture}}
    \centering
    \begin{tabular}{lclclclclc}
    \\
    \toprule
    
    Layer & Feature Maps & Size & Kernel Size & Stride & Activation \\
    \cmidrule(r){1-6}
    Input & 3 & 32x32 & - & - & - \\
    Conv1 & 6 & 28x28 & 5x5 & 1 & ReLU1 \\
    MaxPool & 6 & 14x14 & 2x2 & 2 & - \\
    Conv2 & 16 & 10x10 & 5x5 & 1 & ReLU2 \\ 
    MaxPool & 16 & 5x5 & 2x2 & 2 & - \\ 
    Conv3 & 120 & 1x1 & 5x5 & 1 & ReLU3 \\ 
    FC1 & - & 84 & - & - & ReLU4 \\
    FC2 & - & 10 & - & - & softmax \\
    \bottomrule
    \end{tabular}
    \label{table:lenet_architecture}
\end{table}

\begin{table}[h]
    \caption{\textbf{VGG-9 architecture}}
    \centering
    \begin{tabular}{lclclclclclc}
    \\
    \toprule
    
    Layer & Feature Maps & Size & Kernel Size & Padding & Stride & Activation \\
    \cmidrule(r){1-7}
    Input & 3 & 32x32 & - & - & - & - \\
    Conv1 & 64 & 32x32 & 3x3 & 1 & 1 & BN+ReLU \\
    MaxPool & 64 & 16x16 & 2x2 & - & 2 & - \\
    Conv2 & 128 & 16x16 & 3x3 & 1 & 1 & BN+ReLU \\ 
    MaxPool & 128 & 8x8 & 2x2 & - & 2 & - \\ 
    Conv3 & 256 & 8x8 & 3x3 & 1 & 1 & BN+ReLU \\ 
    Conv4 & 256 & 8x8 & 3x3 & 1 & 1 & BN+ReLU \\
    MaxPool & 256 & 4x4 & 2x2 & - & 2 & - \\ 
    Conv5 & 512 & 4x4 & 3x3 & 1 & 1 & BN+ReLU \\
    Conv6 & 512 & 4x4 & 3x3 & 1 & 1 & BN+ReLU \\
    MaxPool & 512 & 2x2 & 2x2 & - & 2 & - \\
    Conv7 & 512 & 2x2 & 3x3 & 1 & 1 & BN+ReLU \\
    Conv8 & 512 & 2x2 & 3x3 & 1 & 1 & BN+ReLU \\
    MaxPool & 512 & 1x1 & 2x2 & - & 2 & - \\
    FC1 & - & 100 & - & - & - & softmax \\
    \bottomrule
    \end{tabular}
    \label{table:vgg_architecture}
\end{table}

Ensembles of up to 7 VGG-9 networks were trained on the CIFAR-100 data set \cite{krizhevsky_2009}. This network architecture is similar to VGG-11 described in \cite{simonyan_2015}, but here only one fully-connected layer at the top of the network instead of 3 (Figure \ref{fig:architecture}b and Table \ref{table:vgg_architecture}). Networks were trained with 8 different splits of train and test data and random initialization of the weights. The networks' parameters were initialized with the Xavier initialization procedure at the beginning of training. Training was performed using mini-batch stochastic gradient descent with batch size of 256, momentum of 0.9, and weight decay 0.0005. Networks were trained for 150 epochs reaching prefect accuracy on the train set (see below) with cosine annealing schedule (see above) and $\eta_0 = 0.01$. Due to the computational cost of training the networks over the entire $\bar{\beta} = \beta \cdot N$ range, we focused on coupling values with $\bar{\beta} \leq 0$ values for each ensemble size.

In all cases, the data set was split randomly for each training session, into a train set of 50,000 samples and a test set of 10,000 samples. Pixel intensities of images were normalized to values between 0 and 1 as a pre-processing stage; for data augmentation we used random crops and random horizontal flips \cite{krizhevsky_2012}.

Training and analysis code was written using the PyTorch library \cite{pytorch_2019} and was executed on an IBM Spectrum LSF cluster with 5 machines each with a single NVIDIA Tesla-P100 GPU. Training of a single LeNet-5 architecture on CIFAR-10 on a single GPU lasted approximately 31 minutes. 

\newpage 

\subsection*{Ensemble combination methods}
\label{EnsembleCombination}
\begin{figure} [h]
  \centering
  \includegraphics[width=\textwidth]{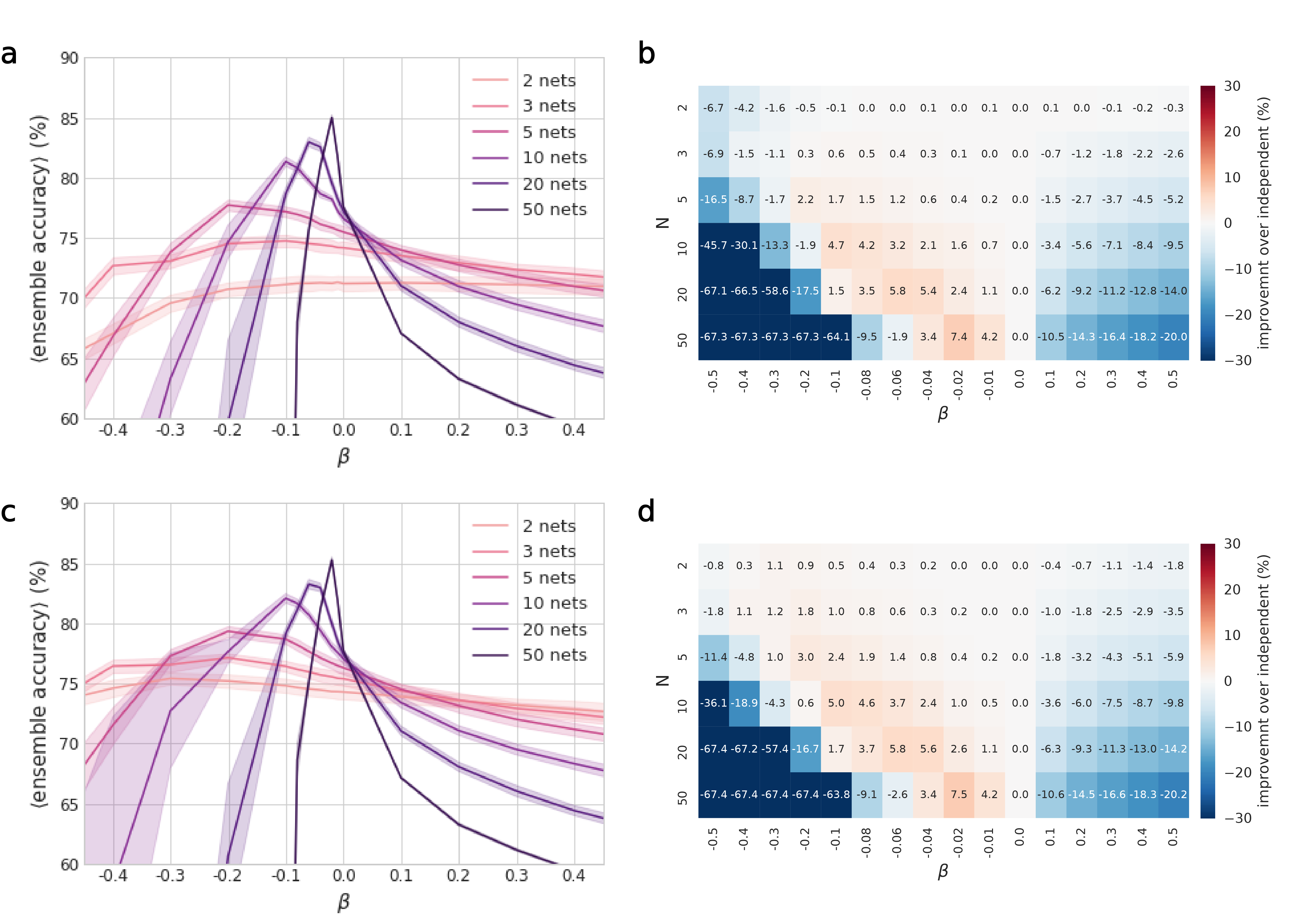}
  \caption{{\bf Optimal negative coupling of co-learning networks for different ensembles size and combination methods.} {\bf a.} Average accuracy of ensembles merged using the majority voting combination method is shown as a function of the coupling coefficient for different ensemble sizes. Funnels around the lines show the standard deviation of accuracy values over 30 runs with different train-test splits and random initialization. ($^*$ For $N=50$ the results here were obtained from 10 repeats only). {\bf b.} Heatmap of the average difference in accuracy between ensembles of coupled networks and ensembles of independently trained networks. {\bf c, d.} same as panels a, b for ensembles merged with the geometric mean combination.}
  \label{fig:other_comb_methods}
\end{figure}

The classification values of the ensembles were computed by linear uniform combination of the individual networks, as described in the main text. We also explored majority voting combination scheme, where the class label predicted by the ensemble is the most common among the set of individual networks, and the geometric mean of the individual networks' outputs $p_i(y|x)$, in which the predicted label by the ensemble $\tilde{y}$ is given by: $\argmax_y{\frac{1}{Z}\prod_{i=1}^N p_i(y|x)}$, where $Z$ is a normalization factor. We repeated the analysis of figure 3 in the main text using these two combination methods, and estimated the ensembles' accuracy for different ensemble size and coupling coefficient values. Figure \ref{fig:other_comb_methods} shows that in all cases the results were qualitatively similar to the results presented in the main text.

\newpage \subsection*{Performance of ensembles of co-learning VGG-9 networks on CIFAR-100}
\label{VGGresults}
\begin{figure} [h]
  \centering
  \includegraphics[width=\textwidth]{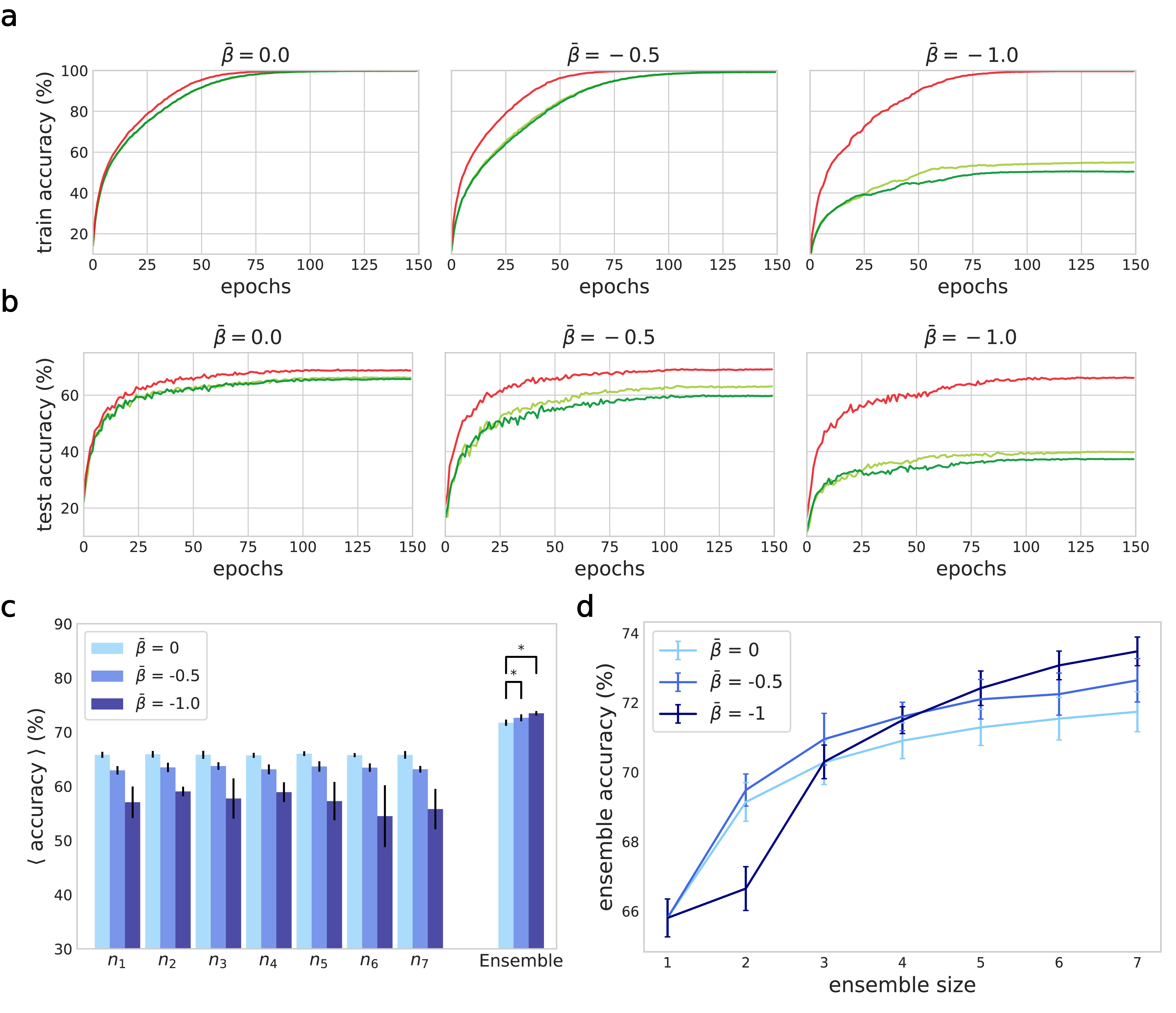}
  \caption{{\bf a.} Example of the training curves of an ensemble of size $N=2$ and its individual networks (VGG-9 trained on CIFAR-100). {\bf b.} Example of the testing curves, same as panel a.
  {\bf c.} The average performance over test data of co-learning ensemble of $N=7$ VGG-9 networks, trained and tested on the CIFAR-100 data set, for negatively coupled networks and independent ones. Error bars show the standard deviation values over the 8 runs. Negatively coupled ensembles outperform independent ensembles for $\bar{\beta} = -0.5$ (Wilcoxon signed-ranks test with Bonferroni correction, $T^+=0, p < 0.05$) and $\bar{\beta} = -1$ (Wilcoxon signed-ranks test with Bonferroni correction, $T^+=0, p < 0.05$). {\bf d.} Comparison of the average test performance of ensembles for different size and coupling values. Negatively coupled ensembles had better performance than independent ones; notably the optimal $\beta<0$ value varies with $N$, as reflected by the crossing of curves for the two negative values. Error bars show the standard deviation values over the 8 runs.}
  \label{fig:vgg_results}
\end{figure}

Figure \ref{fig:vgg_results}a and Figure \ref{fig:vgg_results}b show a typical example of the learning curves for one ensemble of two networks. Here training results in saturation of the ensemble, while individual networks converge to solutions that are farther away as $\bar\beta$ becomes more negative. Figure \ref{fig:vgg_results}c shows the performance of individual networks and of ensembles for different coupling values and ensembles sizes of VGG networks. The performance of ensembles of negatively coupled networks was significantly higher than the performance of independent networks, and the gap grows with ensemble's size (Figure \ref{fig:vgg_results}d).

\newpage 

\subsection*{The effect of ensemble coupling on sparseness of neural activation, synaptic weights, and networks' confidence in labeling}
\label{Sparseness}
Networks that co-learned with different coupling between them, converged to individual networks that had different average activation of units as well as distributions of weights (synapses) between layers. An example for one layer was presented in the main text, and here we show this across all weights and activation layers in the networks. Table \ref{table:sparsness} lists the sparsity of activation for different coupling regimes by (1) the fraction of non-activated neurons and (2) the mean activation in each layer of the network. Networks in negatively coupled ensembles exhibited sparse activation patterns in all layers. 

\begin{table}[h]
    \caption{Individual networks' activation in an ensemble of size $N=2$ under different coupling regimes averaged over 30 train-test splits with random initialization. Negatively coupled ensembles have higher fraction of inactive units but higher average firing rate across all layers.}
    \label{table:sparsness}
    \centering
  \begin{tabular}{lllllll}
  \\
    \toprule
    
    Layer  & 
    \multicolumn{3}{c}{Mean non-active neurons (\%) } &
    \multicolumn{3}{c}{Mean activation (a.u)} \\
    \cmidrule(r){2-7}
    \
    & -0.4 & 0 & 0.4 & -0.4 & 0 & 0.4 \\
    \cmidrule(r){2-4} 
    \cmidrule(r){5-7} 
    ReLU1 & \textbf{$61\pm 0.02$} & $49\pm 0.02$ & $48\pm 0.03$ & \textbf{$0.36\pm 0.02$} & $0.29\pm 0.02 $ & $0.28\pm 0.02$   \\
    ReLU2 & \textbf{$62\pm 0.02$} & $49\pm 0.03$ & $46\pm 0.01$  & \textbf{$0.88\pm 0.08$} & $0.55\pm 0.02$ & $0.55\pm 0.03$ \\
    ReLU3 & \textbf{$70\pm 0.01$} & $57\pm 0.01$ & $55\pm 0.00$ & \textbf{$1.11\pm 0.14$} & $0.52\pm 0.02$ & $0.52\pm 0.02$\\ 
    ReLU4 & \textbf{$70\pm 0.03$} & $54\pm 0.02$ & $54\pm 0.01$ & \textbf{$0.68\pm 0.07$} & $0.32\pm 0.01$ & $0.30\pm 0.02$\\
    \bottomrule
  \end{tabular}
\end{table}

\begin{figure}[!htbp]
  \centering
  \includegraphics[width=\textwidth]{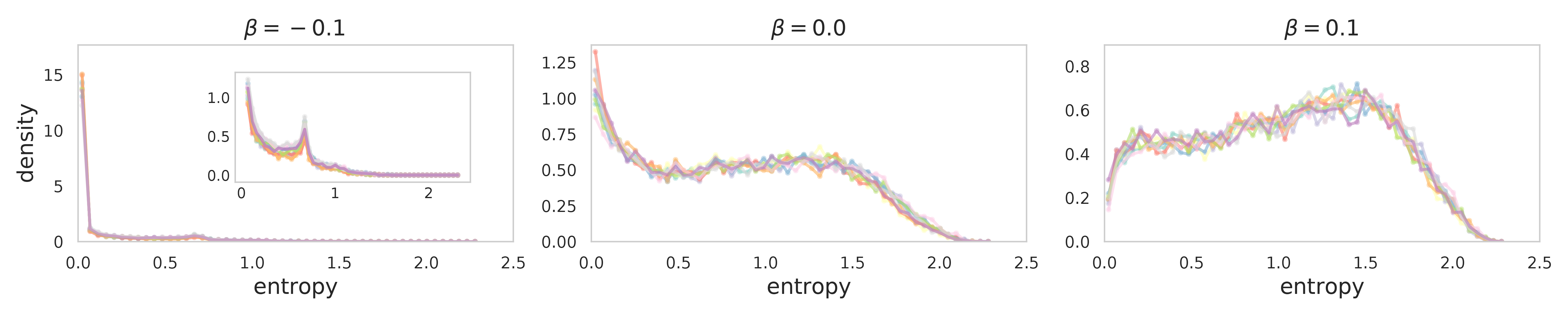}
  \caption{Histograms of entropy values of the classifications of samples by the individual networks in a representative ensemble $H[p_i(y|x)]$, for $N=10$. Lines of different color denote different individual networks in the ensemble, which show very similar entropy distributions even in the case of $\beta<0$. Inset in left panel shows the histogram without the first bin. Negatively coupled networks have high `confidence' in their assessments, reflected by the large fraction of samples with low entropy.}
  \label{fig:entropy_5nets}
\end{figure}

Moreover, negatively coupled networks utilized a wider range of weight values across all layers (Table \ref{table:weights_std}). Extending the analysis of figure 5 in the main text, Figure \ref{fig:entropy_5nets} shows the distribution of entropy values of each of the individual networks in an ensemble of $N=10$ networks, over the samples, $H[p_i(y|x)]$. All the networks in negatively coupled ensembles had a sharp peak close to 0, reflecting that they all became more confident in their classification of samples.

\begin{table}[h]
  \caption{Standard deviation of the weights of individual networks in an ensemble of size $N=2$ under different coupling regimes, averaged over 30 train-test splits with random initialization. 
  }
  \label{sample-table}
  \centering
  \begin{tabular}{llll}
    \\
    \toprule
    
    Layer  & 
    \multicolumn{3}{c}{Parameters Standard Deviation} \\
    \cmidrule(r){2-4}
    \
    & $\beta = -0.4$ & $\beta=0$ & $\beta=0.4$  \\
    \cmidrule(r){2-4} 
    Conv1 & \textbf{$0.45\pm 0.06$} & $0.31\pm 0.01 $ & $0.30\pm 0.01$   \\
    Conv2 & \textbf{$0.21\pm 0.01$} & $0.14\pm 0.00$ & $0.13\pm 0.00$ \\
    Conv3 &  \textbf{$0.08\pm 0.01$} & $0.05\pm 0.00$ & $0.05\pm 0.00$\\ 
    FC1 & \textbf{$0.10\pm 0.00$} & $0.07\pm 0.00$ & $0.07\pm 0.02$\\
    FC2 & \textbf{$0.23\pm 0.01$} & $0.17\pm 0.00$ & $0.16\pm 0.00$\\
    \bottomrule
  \end{tabular}
  \label{table:weights_std}
\end{table}

\newpage 
\subsection*{Expanding ensembles and boosting their performance}
\label{Boosting}
We compared the performance of ensembles of 10 networks trained with their optimal negative $\beta$, to ensembles that combined these 10 networks with 5 additional networks that were trained in different ways. We found that adding 5 independently trained networks or adding 5 networks that were negatively coupled (between themselves and to the original 10 networks), while the first 10 networks were excluded from the optimization process (i.e. were `frozen') did not improve the ensemble's performance (noted as `add-freeze' with $\beta = 0$ or $\beta=-0.07$ in Table \ref{table:addingnets}). When we allowed the existing 10 networks to be optimized together with the new coupled networks, the ensemble's results improved but may be explained by the extra training of the original ensemble - training an ensemble of 15 networks from scratch for 300 epochs reached higher performance. Thus, simple expansions of negatively coupled ensembles require adaptive or learning based expansions of ensembles to be on par with training novel ensembles. We experimented with a simple boosting of the ensemble by training additional 5 independent networks on a subset of the train set. This subset consisted of samples that were close to the decision border: these were the samples for which 5 of the networks of the original ensemble voted correctly and the other 5 voted for another label. This simple boosting-like scheme did not improve the ensemble's performance ($83.32\pm 0.30$ accuracy). We also tried to train the new networks with a broader subset, composed of the training samples for which  $5-m$ to $5+m$ (with $m=1...5$) networks of the the original ensemble voted correctly. This did not improve the ensemble performance either and resulted similar performance for all $m$. While preliminary, these attempts attest to the need for better approach for boosting of ensembles, as in all cases retraining 15 negatively coupled networks from scratch gave the best results.

\begin{table}[H]
    \caption{Different ways to expand a trained ensemble of size 10 with additional 5 new networks as described in the text. Retraining 15 negatively coupled networks from scratch with a large budget was a significantly better strategy.}
    \centering
    \begin{tabular}{lclclc}
    \\
    \toprule
    
    Training Method  & $\beta$ & Ensemble Accuracy (\%) \\
    \cmidrule(r){1-3}
    10 networks ensemble & -0.07 & $83.65\pm 0.38$ \\
    add-freeze-retrain & 0 &  $83.43\pm 0.34$ \\ 
    add-freeze-retrain & -0.07 & $83.74\pm 0.34$ \\
    add-retrain & -0.07 & $\mathbf{84.97\pm 0.33}$    \\
    15 networks ensemble (150 epochs) & -0.07 & $84.29\pm 0.25$ \\
    15 networks ensemble (300 epochs) & -0.07 & $\mathbf{85.65\pm 0.25}$ \\
    \bottomrule
    \end{tabular}
    \label{table:addingnets}
\end{table}

\end{document}